\documentclass[10pt,twocolumn,letterpaper]{article}

\usepackage[pagenumbers]{cvpr}              %

\newcommand{\modelname}{{\color{black}ExpertAF}}

\usepackage{array}
\usepackage{colortbl}
\newcommand{\PreserveBackslash}[1]{\let\temp=\\#1\let\\=\temp}
\newcolumntype{C}[1]{>{\PreserveBackslash\centering}p{#1}}
\newcolumntype{L}[1]{>{\PreserveBackslash\raggedright}p{#1}}

\DeclareMathOperator*{\argmin}{\arg\!\min}

\definecolor{Gray}{gray}{0.95}
\definecolor{DarkGray}{gray}{0.55}
\definecolor{LightGray}{gray}{0.95}

\usepackage{tcolorbox}
\tcbuselibrary{breakable} %

\usepackage[framemethod=TikZ]{mdframed} %
\newmdenv[
  backgroundcolor=LightGray,
  linewidth=1pt,
  linecolor=Gray,
  roundcorner=7pt
]{myframe}

\definecolor{cvprblue}{rgb}{0.21,0.49,0.74}
\usepackage[pagebackref,breaklinks,colorlinks,allcolors=cvprblue]{hyperref}

\def\paperID{9484} %
\def\confName{CVPR}
\def\confYear{2025}

\title{\modelname: Expert Actionable Feedback from Video}

\author{Kumar Ashutosh$^{1,2}$, Tushar Nagarajan$^2$, Georgios Pavlakos$^1$,
Kris Kitani$^{2,3}$, Kristen Grauman$^{1,2}$\\
$^1$~University of Texas at Austin, $^2$~FAIR Meta, $^3$~Carnegie Mellon University\\
}

\begin{document}
\maketitle
\begin{abstract}

Feedback is essential for learning a new skill or improving one's %
current skill-level. 
However, current methods for skill-assessment from video only provide scores or compare demonstrations, leaving the burden of knowing what to do differently on the user.
We introduce a novel method to generate \emph{actionable feedback} (AF) from video of a person doing a physical activity, such as basketball or soccer.  Our method takes a video demonstration and its accompanying 3D body pose and generates (1) free-form expert commentary describing what the person is doing well and what they could improve, and (2)
a visual expert demonstration
that incorporates the required corrections. %
We show how to leverage 
Ego-Exo4D's~\cite{egoexo4d} videos of skilled activity and expert commentary together
with a strong language model to create a weakly-supervised training dataset for this task, and we devise a multimodal video-language model to infer coaching feedback.
Our method is able to reason across multi-modal input combinations to output full-spectrum, actionable coaching---expert commentary, expert video retrieval, and  
expert pose generation---outperforming strong vision-language models on both established metrics and human preference studies. 
\renewcommand{\thefootnote}{}\footnotetext{Project page: \href{https://vision.cs.utexas.edu/projects/ExpertAF/}{https://vision.cs.utexas.edu/projects/ExpertAF/}} %
\end{abstract}

\begin{figure}[t]
    \centering
    \includegraphics[width=0.5\textwidth]{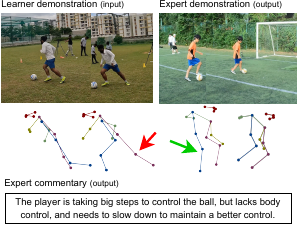}
    \vspace{-0.5cm}
    \caption{\textbf{An example of expert feedback.} When a player is dribbling the ball fast, they tend to lose control (top left). Our proposed method provides an expert commentary to the learner suggesting improvements (bottom). The method also provides an expert demonstration that shows the desired correction, %
    where the player is maintaining smaller steps and body control (top right). 
    }
    \label{fig:teaser}
    \vspace{-0.5cm}
\end{figure}    
\section{Introduction}

An abundance of instructional ``how-to'' videos on the internet enables skill-learning by observing expert demonstrations. Instructional videos cover %
many %
skills one might want to learn---cooking, DIY, sports, crafts, and hobbies. Recent work leverages such videos %
to %
assist human~\cite{task_graph,howto100m,detours} and robot~\cite{r3m} skill learning.  Acquiring skills from video is especially appealing for equitable learning: access to video is generally less costly and more widely available than face-to-face access to an expert coach.

Despite the myriad of videos and tutorials, %
people prefer to learn in a feedback loop: getting to know 
their mistakes, finding the improvements they need to do and correcting those.
For that reason, coaching-based iterative training is more effective than self-learning \cite{self-training-vs-coaching}. A good coach has three components: finding mistakes, providing verbal corrections and finally, showing visual demonstrations. For example, suppose a person learning to dribble a soccer ball takes big steps that decrease body control. The coach should identify this mistake, provide verbal feedback, and %
show the correct dribbling technique. %
See Figure~\ref{fig:teaser}.

At present, a person watching a how-to video has to identify mistakes themselves and attempt to correct them. Pinpointing the exact mistake is itself challenging for beginners---particularly when learning \emph{physical} skills, like sports and dance, where subtle movements and positioning of the body are critical to success. %

What role could AI play in overcoming this gap?  Current work on proficiency understanding from video focuses on \emph{scoring} a demonstration \cite{egoexo4d,skating-eval,gordon1995automated,baller-gedas}, thus addressing only one component of coaching. There is no method that can provide actionable verbal and visual feedback on a learner's demonstration.
Doing so is technically challenging since it requires the method to 
(a) understand the user's activity, (b) detect mistakes in the execution by comparing the user's activity with \emph{correct} %
examples and finally, (c) suggest edits with respect to the better 
way of doing the activity. 
In short, the output should be \emph{coach-like}---specific actionable feedback that will improve a learner's proficiency.

We propose \modelname: a novel method to provide expert actionable feedback on a user's physical activity. \modelname's actionable feedback has two forms: language-based commentary and visual demonstration.  The free-form expert commentary output describes the mistake and what should be corrected, e.g., \emph{``take smaller steps and slow down to maintain control''}, while the expert visual demonstration output shows the correct way of doing it.  %

Both components introduce unique technical challenges.
For the former, the task of generating free-form text feedback %
is distinct from %
standard image and video captioning, as the model needs to identify the %
mistake and actions that would correct it---rather than %
simply describe the observed pose \cite{posescript} or activity~\cite{captioning-1,captioning-2,captioning-3,captioning-4,captioning-5}.
Similarly, for the latter, the task of retrieving (or generating) a demonstration that \emph{corrects} a specific mistake %
is distinct from
similarity-based retrieval~\cite{mil-nce,videoclip,hiervl,egovlp,internvideo2} or open-ended generation~\cite{posegpt,sora}.
We hypothesize---and experimentally validate---that 
personalized, video-conditioned coaching
is more %
accurate than simply returning %
a global expert execution because there are many right ways of doing an activity, e.g., \emph{penalty shots} in soccer, and hence different ways to fix what is wrong in a given execution. 
Overall, the model must understand both what is being done \emph{and} what actionable tweaks would make it better.

We are the first to address video-based coaching %
with actionable feedback.  We develop the first model to 
generate free-form text feedback on a user's activity (captured in video and 3D pose sequences) as well as a %
video demonstrating how to improve, by %
building on a recent vision-language model \cite{llava}. %
There is currently no dataset for this challenging task. Therefore, we augment Ego-Exo4D~\cite{egoexo4d}, an existing dataset with video, commentary, and poses, and show how to use a strong large language model %
to create a weakly-supervised training set %
consisting of paired learner-expert demonstrations, along with expert textual commentary that relates the two. 
We also obtain a gold-standard
manually labeled test set for rigorous evaluation.

We %
validate \modelname~%
on three diverse physical scenarios: %
soccer, basketball, and rock climbing.  %
Our results show a significant improvement over strong baselines, establishing the first method to provide actionable feedback, including a novel technique for 
expert pose generation. %
Alongside consistent gains in quantitative metrics against the ground truth, we also show our model's promise via direct human evaluation, where \modelname~outperforms off-the-shelf video models by as much as 3$\times$. %

\section{Related work}

\textbf{Skill learning from videos.} Instructional video datasets like HowTo100M \cite{howto100m}, COIN \cite{coin}, and CrossTask~\cite{crosstask}  enable procedural understanding through step recognition \cite{video-distant,task_graph,paprika}, procedure planning \cite{procedure-learning-fei-fei-li,procedure2,procedure3}, task-graph discovery \cite{task_graph,paprika,graph2vid,egoexo4d}, action anticipation \cite{avt,rulstm,intention,whenwillyoudowhat,gao2017red}, and alignment detection \cite{vnd,tan}. %
Capitalizing on their instructional nature,
recent work learns robot policies from these videos \cite{dexvip,roboclip}, uses external knowledge-bases like WikiHow to ground the steps of a procedural task%
~\cite{vina,video-distant,paprika}, and explores new ways to navigate %
between multiple demonstrations \cite{detours}. 
Despite their scale, how-to videos in these datasets~\cite{howto100m,coin,crosstask} are often created by domain experts and hence lack mistakes or suboptimal executions.  In contrast, 
Ego-Exo4D \cite{egoexo4d} %
contains multiple executions of various common scenarios by people with varied skill levels---from novice to late expert. Moreover, it contains experts' feedback on those actions. 
We show how to augment this data via large language models to support our new idea for video-based coaching.

\textbf{3D body pose for activity understanding.} 3D body pose is %
crucial 
for human activity understanding, capturing a person's stance and motion.
Supported by valuable datasets like Humans3.6M \cite{human36m} and NTU RGB-D \cite{ntu-rgbd,ntu-rgbd-120}, 
recent methods improve body pose understanding~\cite{motionbert,pose-sota-1,pose-sota-2,pose-sota-3,pose-sota-4,pct}.
Beyond action classification, pose is also crucial for %
markerless motion capture of sports~\cite{sportscap,poseforrunning} and interacting in augmented reality \cite{avatar,poseforar}.
While our \modelname ~leverages pose as an important signal of human activity, unlike prior work we
interface video, pose, and free-form text to generate expert feedback and an expert demonstration video. %
Our work is also different from generating~\cite{chatpose} or modifying~\cite{posefix} 
body pose from text (e.g.,\emph{``raise your left arm''}), as
our task is to understand a potentially incorrect pose and provide feedback. Understanding these minor differences between incorrect and correct poses %
offers unique challenges that are addressed by our proposed learning scheme.  Our work is orthogonal to methods that improve pose estimation itself; any future improvements on that front would only benefit our approach.

\textbf{Skill assessment and coaching.} Prior work explores skill assessment %
for a variety of tasks, particularly sports. Most methods pose skill-assessment as a score-prediction task, i.e., for skating \cite{skating-eval}, gymnastics \cite{gordon1995automated}, basketball \cite{baller-gedas}, or multiple sports \cite{stl-vs-mtl}. Since absolute scoring can be ambiguous and error-prone, %
recent work~\cite{quality-distribution-learning} explores
uncertainty-aware score distributions.  Instead of explicitly scoring a demonstration, other work %
determines which of two basketball players are better~\cite{baller-gedas}, uses group-aware contrastive regression to learn the relative quality of demonstrations~\cite{group-aware-regression}, or generates a full ideal trajectory from the first frame \cite{basketball-planning-gedas}. Fitness-AQA \cite{workout-form-assessment} and Action Quality \cite{interpretable-feedback} provide outputs like \emph{knees inward error}, \emph{shallow squat error}, or an arrow highlight for more localized feedback. There is also research on skill assessment in non-sports domains, like surgery~\cite{skill-surgical,skill-surgical-2} and piano~\cite{skill-piano}. 
All the prior work assumes a fixed taxonomy of errors, and the taxonomy is designed separately for each activity. %
Furthermore, unlike our work, none of the prior work provides feedback akin to a personal coach---which requires  %
not only telling what is wrong, but also expressing how to correct the mistake.

\vspace{-0.1cm}
\section{Method}
\label{sec:method}

We introduce the problem statement in Sec. \ref{sec:problem-statement}, the dataset creation strategy in Sec. \ref{sec:dataset-creation}, the training design in Sec. \ref{sec:training-design}, and  %
implementation details in Sec. \ref{sec:implementation-details}. %

\subsection{Problem statement}
\label{sec:problem-statement}

Consider a dataset $\mathcal{E} = \{(\mathcal{V}, T, \bar{\mathcal{V}})\}$ where each $\mathcal{V}_i$ is a video demonstration, and $T_i$ is a free-form text commentary critiquing the activity in the video, e.g., \emph{``take smaller steps and slow down to maintain control.''} $\bar{\mathcal{V}}$ is a related %
video demonstration but without the error mentioned in $T$, e.g., the 
player takes smaller steps with better control. See Fig. \ref{fig:teaser}. Each demonstration $\mathcal{V} = \{V, P, S\}$ consists of three parts: $V$ is the RGB video, and $P$ is the 3D pose sequence of the participant with skill-level $S$ in the video.
The pose representation $P \in \mathbb{R}^{n \times d \times 3}$ contains $n$ frames of %
3D positions for $d=17$ body joints, consistent with MS-COCO~\cite{coco}. $S$ is the participant's skill-level for an activity, i.e., %
\emph{novice}, \emph{early expert}, \emph{intermediate expert}, or \emph{late expert}, and will be used later in creating the dataset.
We assume there is one active person doing a physical activity in any given video who is the subject of \modelname.

The goal in this work is to provide feedback on a given learner's video %
containing a physical scenario. The feedback is %
an expert demonstration $\bar{\mathcal{V}}$ and %
an expert commentary $T$, tailored to $\mathcal{V}$. Formally, we want to learn a mapping $\mathcal{V}~\rightarrow~(\bar{\mathcal{V}}, T)$. While it is possible to learn these mappings separately (i.e., $\mathcal{V} \rightarrow \bar{\mathcal{V}}$ and $\mathcal{V} \rightarrow T$), the expert commentary and demonstration are tightly related and provide important contextual information for generating each other. We therefore treat this as an autoregressive generation problem where during training, we use both $\mathcal{V}$ and $T$ to generate $\bar{\mathcal{V}}$ (or $\mathcal{V}$ and $\bar{\mathcal{V}}$ to generate $T$), while during testing, we  drop the extra context information to generate $\bar{\mathcal{V}}$ and %
$T$ directly.

Mathematically, consider $\mathcal{F}: (\mathcal{V}, \bar{\mathcal{V}}, T) \rightarrow \mathbb{R}$ as the 
autoregressive function that jointly learns from the learner demonstration $\mathcal{V}$, the expert demonstration $\bar{\mathcal{V}}$, and the expert commentary $T$. The output $\in \mathbb{R}$ is the loss that we aim to minimize.  We use the same unified $\mathcal{F}$ for the joint training (detailed in Sec.~\ref{sec:training-design}) and inference, as follows. We use $\mathcal{F}'$ when the autoregressive model $\mathcal{F}$ is used for output token generation, i.e., text or pose tokens. $\mathcal{F}'$ can take in any input used during training and generate the remaining one---typical for autoregressive models \cite{vaswani2017attention}. %

\textbf{Expert commentary generation.} %
At inference time, to generate an expert commentary $T$, %
we only have a learner demonstration, thus we mask out the expert demonstration and output %
the expert commentary:
\begin{align}
\label{eq:comm-gen}
    \hat{T} = \mathcal{F}_t (\mathcal{V}) = \mathcal{F}'(\mathcal{V}, \varnothing)
\end{align}
where $\mathcal{F}_t$ denotes using the model to generate commentary.

\textbf{Expert demonstration generation.} %
Next, we obtain the expert demonstration in two %
forms---retrieving a correct execution (video and pose) and generating a pose sequence. These two output formats offer complementary information to understand the actionable feedback. %
While a video exemplar is helpful for many learner mistakes, pose generation is useful in the absence of a correct expert demonstration in the retrieval set---allowing our model to generalize its coaching beyond the set of discrete expert videos.  %
We leave expert video generation as a future work. %

Denoting $\mathcal{F}_r$ and $\mathcal{F}_g$ as the expert demonstration retrieval and expert pose generation functions, respectively, we have: 
\begin{align}
    \bar{\mathcal{V}} &= \mathcal{F}_r(\mathcal{V}) = \argmin_{\forall \mathcal{V}' \in \mathcal{E}} \mathcal{F}(\mathcal{V}, \mathcal{V}', \hat{T}) \\
    P &= \mathcal{F}_g(\mathcal{V}) = \mathcal{F}'(\mathcal{V}, \hat{T})
\end{align}
where $\hat{T}$ is the output from Eq. $\ref{eq:comm-gen}$ above, and $\mathcal{F}_r$ and $\mathcal{F}_g$ denote using the model to retrieve or generate the expert demonstration, respectively.

In summary, the training step $\mathcal{F}$ uses %
tuples $\mathcal{V}, \bar{\mathcal{V}}, T$ in a unified way, described in Sec.~\ref{sec:training-design}, whereas, at inference, all the three functions $\mathcal{F}_t$, $\mathcal{F}_r$ and $\mathcal{F}_g$ use only $\mathcal{V}$.

\begin{figure*}[t]
    \centering
    \includegraphics[width=\textwidth]{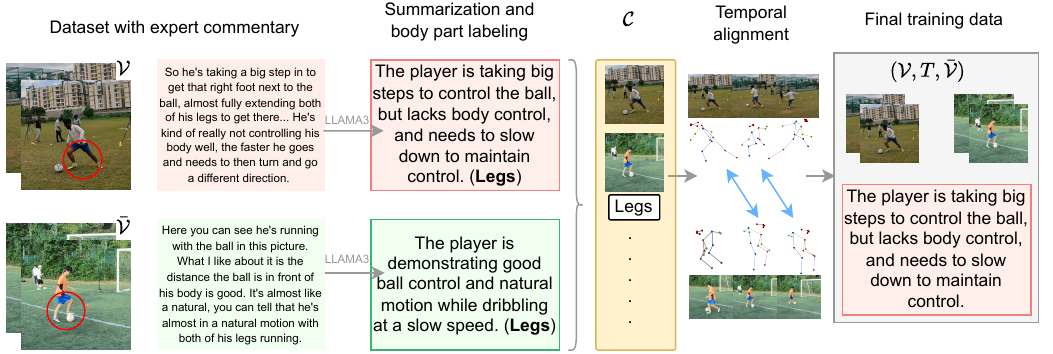}
    \caption{\textbf{Overview of the dataset creation.} We first summarize the human-provided expert commentary~\cite{egoexo4d} in one sentence using an LLM, and then map it to a body region and correct (green) or incorrect (red) execution label. We then choose incorrect-correct pairs for the same body region to obtain $\mathcal{C}$. Finally, we %
    choose pairs with minimum temporal alignment loss to obtain the training data. Best in zoom.}
    \label{fig:dataset-overview}
    \vspace{-0.5cm}
\end{figure*}

\subsection{Forming the expert feedback dataset}
\label{sec:dataset-creation}

To learn the desired functions, we need to obtain pairs of videos $(\mathcal{V}, \bar{\mathcal{V}})$ where there is an error in the demonstration in the first video %
that is
corrected in the second video, along with the expert commentary $T$ about $\mathcal{V}$.
Ego-Exo4D \cite{egoexo4d} offers a great starting point for %
our setup. It contains ego-exo videos, %
extracted 3D pose sequences, %
and time-stamped commentary by experts (e.g., professional soccer coaches) on the demonstrations in the video. The experts watched the entire video and stopped each time they saw something to critique or compliment, offering free-form spoken commentary.  In total, this led to 117,812 sentences across 221 hours of video, with most videos commentated by 2-5 unique experts~\cite{egoexo4d}.
See examples in Figure~\ref{fig:dataset-overview} (left) and details in Sec. \ref{sec:implementation-details} and Supp. 

We propose to automatically augment this data in two ways to enable coaching.  
First, we seek (pseudo-)annotations of whether a given %
commentary statement %
describes a needed improvement or %
applauds a correct execution. Second, we localize each piece of feedback %
on a body region, e.g., \emph{incorrect hand stretch} vs.~\emph{wrong legs movement}. These distinctions are crucial to generate feedback and show corrections. %
To this end, we create a weakly-supervised training dataset %
consisting of tuples $(\mathcal{V}, T, \bar{\mathcal{V}})$ from Ego-Exo4D \cite{egoexo4d},  as follows.  %

\textbf{Expert commentary classification and body localization.} The  commentary in \cite{egoexo4d} is obtained from voice recordings of the experts converted to text with ASR. Most of the samples contain extra comments like \emph{``oh, I will give this a five out of ten''} and \emph{``that's how I would do it too''}. Additionally, as discussed above, they lack positive and negative labels and do not explicitly indicate the body region involved (e.g., \emph{legs}, \emph{arms}). Thus, we preprocess the expert commentary for three things---making the commentary concise to extract the improvable feedback, marking which body region the feedback is about, and %
marking whether the feedback states the need for improvement or not. 

Since these %
are all addressable with
text reasoning, we use a large language model (LLM) to provide the desired answers. We use Llama3-70B~\cite{llama3modelcard}, a recent open-sourced model that performs well in current benchmarks.
In essence, given a commentary $T$, the language model $\mathcal{L}$ %
responds to the prompt: \emph{``Given an expert's commentary, summarize the feedback into a single sentence and also provide which body regions need improvement or are correctly executed...''} (see Supp.). Formally, this yields $\mathcal{L}(T) = (T', (b^1, c^j), ..., (b^s, c^s))$ where $T'$ is the concise summary and $b^i$ is a body region like \emph{head} or \emph{arms} and
$c^i \in \{0, 1, 2\}$ %
are labels representing \emph{needs improvement}, \emph{correct execution}, and \emph{no mention}, respectively. We group the skeleton joints into six pre-defined body regions $b^i$ (details in Supp.).  See red and green boxes in Fig.~\ref{fig:dataset-overview}. %

\textbf{Pairing incorrect and correct executions.} Next, we use the above information to mine pairs of incorrect and correct execution in the dataset. %
Ego-Exo4D also contains metadata about the skill-level $S$ of the demonstrator, broken into four categories (1-4 in increasing expertise) starting
from \emph{novices} who have not performed the activity before to \emph{late experts} who have performed the same activity, e.g., \emph{basketball} for $10+$ years. We sample %
incorrect learner demonstrations from beginners and correct demonstrations from experts from the same activity in Ego-Exo4D, e.g., penalty kicks in soccer or reverse layup in basketball. Even though there can be incorrect executions by experts and correct ones by beginners, errors by experts are likely incomparable to beginners' due to the skill gap.
We use the mapping of body regions to find (in)correct executions referring to the same body region. This results in a collection of video pairs with negative and positive feedback about the same body region, e.g., legs in Fig. \ref{fig:dataset-overview}. Formally, the collection $\mathcal{C}$ is curated as
\begin{displaymath}
    \{(\mathcal{V}_1, T, \mathcal{V}_2)~|~S_1 \in \mathcal{S}^n, S_2 \in \mathcal{S}^{e}, \exists j~s.t.~ c_1^j = 0, c_2^j = 1\}
\end{displaymath}
where $\mathcal{S}^n = $\{novice, early expert\} and $\mathcal{S}^e = $ \{late expert, intermediate expert\}. Using this matching, we obtain a collection of video pairs of incorrect and correct execution (Fig.~\ref{fig:dataset-overview}, yellow box). %

\begin{figure*}
    \centering
    \includegraphics[width=\textwidth]{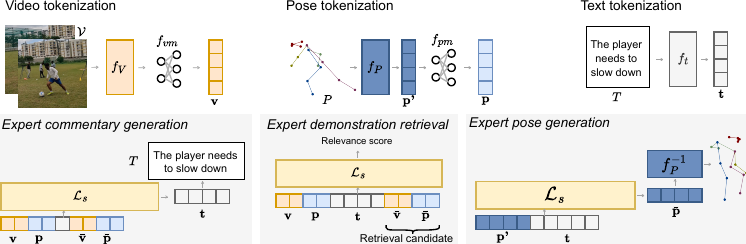}
    \caption{\textbf{Model overview.} We tokenize individual modalities using a modality-specific architecture (top). Once all the modalities are encoded as tokens, we use a %
    large language model to learn expert commentary generation, demonstration retrieval, and pose generation. At inference, the model only takes the learner demonstration video $\mathcal{V}$. See text for details.}
    \label{fig:method}
    \vspace{-0.5cm}
\end{figure*}

\textbf{Temporal alignment and filtering demonstrations.} 
The commentary annotation process in \cite{egoexo4d} allows the experts to pause at any instant and provide their feedback. Thus, the clip $\mathcal{V}_1$ could have the start of a basketball jump shot, whereas the clip $\mathcal{V}_2$ could also have content before the shot and the follow-through. Thus, we temporally align the learner and the expert video in the collection $\mathcal{C}$ to ensure we are capturing the same step of a demonstration.

For all pairs of clips in the collection $\mathcal{C}$, we compare the two poses $P$ to quantify their %
alignment (similarity) using Procrustes-aligned Mean Per Joint Position Error (PA-MPJPE) \cite{4dhumans}, a translation-invariant and body shape-invariant measure.
Note that the video component of the input (Sec.~\ref{sec:results}) is complementary, as it does capture the person's overall movement in the space. %

Denote the commentary timestamp as $t_1$ and $t_2$ for the two videos. We first choose a fixed window around $t_1$, say $[t^a_1, t^b_1]$ such that $t_1^a < t_1 < t_1^b$. Next, we find the corresponding time window $[t^a_2, t^b_2]$ with $t_2^a < t_2 < t_2^b$ in $\mathcal{V}_2$ such that the PA-MPJPE error %
is the minimum. Specifically, %
\begin{align*}
    [t^a_2, t^b_2] = \argmin_{[t^i_2, t^j_2] \in \mathcal{V}_2} \text{PA-MPJPE} \left( \mathcal{V}_1[t^a_1, t^b_1], \mathcal{V}_2[t^i_2, t^j_2] \right)\\
    ~~\text{such that}~~|t^j_2 - t^i_2| = |t^b_1 - t^a_1|.
\end{align*}

We obtain a subset from $\mathcal{C}$ that contains the aligned video segments. %
There can be some video pairs where the demonstration can be very different and hence no appropriate match exists. Therefore, we only keep the top-k pairs with maximum alignment %
for every incorrect execution.

Overall, we obtain a novel dataset $\mathcal{D}$ 
that contains pairs of videos where the first video has an incorrect execution (on some body region) and the second video corrects it, thus obtaining the desired tuple $(\mathcal{V}, T, \bar{\mathcal{V}})$ for training and testing.  See Fig.~\ref{fig:dataset-overview} (right) and examples in Supp.
To ensure fair evaluation, we separately establish a clean gold standard test set, free of potential noise from LLM inference.  The test set is manually verified (see Supp for details).

\subsection{Architecture and training design}
\label{sec:training-design}

Next, we discuss the architecture that encodes the videos $\mathcal{V}$, poses $P$, and text $T$ and enables training the auto-regressive function $\mathcal{F}$. 
The overall idea is to encode all the representations into a text embedding space and use the strong capabilities of recent language models to obtain output text and pose tokens. This approach has been recently used in vision-conditioned language models \cite{llava,llava-med,detours,stepdiff} for image and video captioning. This setup also allows for a unified architecture for various input and output combinations, as opposed to different input streams for individual modalities. %
 Fig.~\ref{fig:method} shows a schematic diagram of the architecture, and each part is explained below.

\textbf{Encoding video as tokens.} %
Each video $\mathcal{V}$ is an ego-exo clip pair.\footnote{The exo view is the one annotated in Ego-Exo4D as having the maximum subject visibility.} The input thus allows
the model to see both close-up hand-object interactions (more visible in the learner's ego view) as well as full-body poses in the scene context (more visible in the observer's exo view).
We use a pre-trained video model 
to extract spatio-temporal features from both videos' frames, followed by a mapper to convert the video features to video tokens. Formally, $\textbf{v} = f_{vm}(f_V(\mathcal{V))}$ where $f_V$ is a standard 
feature extractor (we use InternVideo2~\cite{internvideo2}) and $f_{vm}$ is the visual mapper. See Fig. \ref{fig:method} (top left). %
The mapper is typically a low-parameter model that is trainable, whereas the high-parameter feature extractor is kept frozen. The mapper helps transform the visual mapper to be akin to text tokens---a popular strategy in visual instruction tuning models~\cite{llava,llava-med}.

\textbf{Encoding pose sequences as tokens.}  To encode a pose sequence, we 
use a series of linear layers and MLP-mixer~\cite{pct} to convert a single pose $P \in \mathbb{R}^{d\times 3}$ to an embedding that can be discretized using a codebook, i.e., $\mathbf{p}' = f_P(P)$. The architecture in \cite{pct} also contains a decoder that converts the embeddings back to human poses, i.e., $P = f_P^{-1}(\mathbf{p}')$, which we use in $\mathcal{F}_g$ to generate corrected poses.
Similar to the above, we use a pose mapper to convert the tokens to embeddings. Formally, $\mathbf{p} = f_{pm}(f_P(P)) = f_{pm}(\mathbf{p}')$ is the pose token where $f_{pm}$ is the pose mapper.  See Fig. \ref{fig:method} (top center). We concatenate embeddings from every frame to obtain the representation for the whole sequence. Having $f_{pm}$ allows training with fewer parameters and thus, we use $\mathbf{p}$ for learning $\mathcal{F}_t$ and $\mathcal{F}_r$. However, using $f_{pm}$ for generation $\mathcal{F}_g$ would require adding an inverting function for $f_{pm}$. Thus, we directly add $\mathbf{p}'$ to LLM tokens for pose generation $\mathcal{F}_g$. %
An innovative aspect of our approach is to encode pose as multimodal tokens and train with LLMs, unlike other work regressing the pose parameters from a special pose embedding \cite{chatpose} or using a dedicated pose transformer \cite{t2m-gpt,couple-dance}.

\textbf{Encoding text as tokens.} Text tokenization is the standard process before inputting a text sentence to the LLM \cite{llama2,gpt4}, 
i.e., $\mathbf{t} = f_t(T)$ where $f_t$ is the tokenization function. 
 See Fig. \ref{fig:method} (top right). 

\textbf{Multi-modal sequence prediction.} The previous steps yield multimodal tokens $\mathbf{v}$, $\mathbf{p}$, and $\mathbf{t}$ for video, pose, and text, respectively. Next, we use the strong sequence prediction capabilities of large language models to obtain the desired output tokens based on the sequence of multi-modal inputs \cite{llava,llava-med,detours,stepdiff}. 

For training $\mathcal{F}$ for expert commentary generation, we provide a sequence of learner video and pose tokens ($\mathbf{v}$ and $\mathbf{p}$) and the corrected pose tokens ($\mathbf{\bar{v}}$ and $\mathbf{\bar{p}}$ corresponding to $\bar{\mathcal{V}}$). We ask the model to predict the expert commentary token sequence. For the sequence prediction language model $\mathcal{L}_{s}$, we wish to obtain $\mathbf{t} = \mathcal{L}_s(\mathbf{v}, \mathbf{p}, \mathbf{\bar{v}}, \mathbf{\bar{p}})$. Consequently, the training objective is the standard cross-entropy loss, $\min_{\theta} \left\{ -\log \left( \mathbf{t}~|~\mathbf{v}, \mathbf{p}, \mathbf{\bar{v}}, \mathbf{\bar{p}}; \theta \right) \right\}$,
where $\theta$ are the parameters of the model.  See Fig.~\ref{fig:method} (bottom left). For consistency with the training of these language models, the sequence is formatted to be conversational in nature, e.g., ``\emph{provide an expert's commentary based on this pose sequence:}''. Similarly, for expert demonstration retrieval, we wish to obtain a retrieval candidate $\mathbf{\bar{p}} = \mathcal{L}_s(\mathbf{v}, \mathbf{p}, \mathbf{t})$. Likewise, the training objective with parameters $\gamma$ is $\min_{\gamma} \left\{ -\log \left( \mathbf{\bar{p}}~|~\mathbf{v}, \mathbf{p}, \mathbf{t}; \gamma \right) \right\}$, which we also use as the relevance score for retrieval during inference.
Expert pose generation is trained with the same objective, except we use $\mathbf{p}'$ instead of $\mathbf{p}$ (and obtain $\bar{\mathbf{p}}'$). See Fig. \ref{fig:method} (bottom). Generating pose further requires converting back the pose tokens $\bar{\mathbf{p}}'$ to 3D joints using the pose decoder $f_P^{-1}$.

At inference time, as introduced in Sec. \ref{sec:problem-statement}, we only use $\mathcal{V}$ as the input: $\mathcal{F}_t$ drops the expert demonstration and predicts the expert commentary $\hat{T}$. We use this predicted commentary as input for $\mathcal{F}_r$ and $\mathcal{F}_g$. %

\begin{figure*}[htbp]
    \centering
    \begin{minipage}{0.65\textwidth}
        \centering
        \footnotesize

\begin{tabular}
{L{2.8cm}C{0.75cm}C{0.75cm}C{0.75cm}C{0.75cm}C{0.75cm}C{0.85cm}}
\toprule
& \multicolumn{3}{c}{\textbf{Commentary Gen.}} & \multicolumn{2}{c}{\textbf{Demo Ret.}}  & \multicolumn{1}{c}{\textbf{Pose Gen.}} \\
\cmidrule(lr){2-4} \cmidrule(lr){5-6} \cmidrule(lr){7-7}
Method & B@4 & M & R-L &  R & medR~$\downarrow$ & P~$\downarrow$ \\
\midrule
InternVideo2-NN \cite{internvideo2} & 42.1 & 46.9 & 49.3 & 13.5 & 198 & 161  \\
InternVideo2-FT \cite{internvideo2}&  42.9 & 47.6 & 50.0 & 14.1 & 190 & 157 \\
VideoChat2 \cite{videochat2} & 27.8 & 44.3 & 41.9 & 14.9 & 183 & ---  \\
LLaVA \cite{llava} & 28.5 & 44.1 & 44.2 & 15.0 & 183 & --- \\
LLaVA-FT \cite{llava} & 43.5 & 48.5 & 51.5 & 17.8 & 177 & ---  \\
LLaVA-FT w/ pose \cite{llava} & 43.6 & 48.5 & 51.7 & 18.0 & 172 & 150  \\
PoseScript/Fix \cite{posescript,posefix} & 24.1 & 44.5 & 46.3 & 15.9 & 182 & 182  \\
\rowcolor{Gray}
\midrule
\textbf{\modelname} & \textbf{44.9} & \textbf{49.6} & \textbf{54.6} & \textbf{19.1} & \textbf{158} & \textbf{135}  \\
\textbf{\modelname} w/o video & 44.6 & 49.4 & 54.2 & 18.7 & 161 & 139  \\
\textbf{\modelname} w/o pose & 44.2 & 49.3 & 54.2 & 18.6 & 163 & ---  \\
~~w/o alignment & 42.0 & 48.6 & 51.5 & 16.9 & 180 & 153   \\
~~w/~~global pose & 43.0 & 47.9 & 52.6 & 16.5 & 184 & 150   \\
\midrule
\textbf{\modelname} w/ full-sup & \textcolor{DarkGray}{\textbf{45.8}} & \textcolor{DarkGray}{\textbf{50.9}} & \textcolor{DarkGray}{\textbf{55.7}}  & \textcolor{DarkGray}{\textbf{22.5}} & \textcolor{DarkGray}{\textbf{146}}  & \textcolor{DarkGray}{\textbf{131}}    \\
\bottomrule
\end{tabular}

    \end{minipage}%
    \hfill
    \begin{minipage}{0.35\textwidth}
        \centering
        \includegraphics[width=\textwidth]{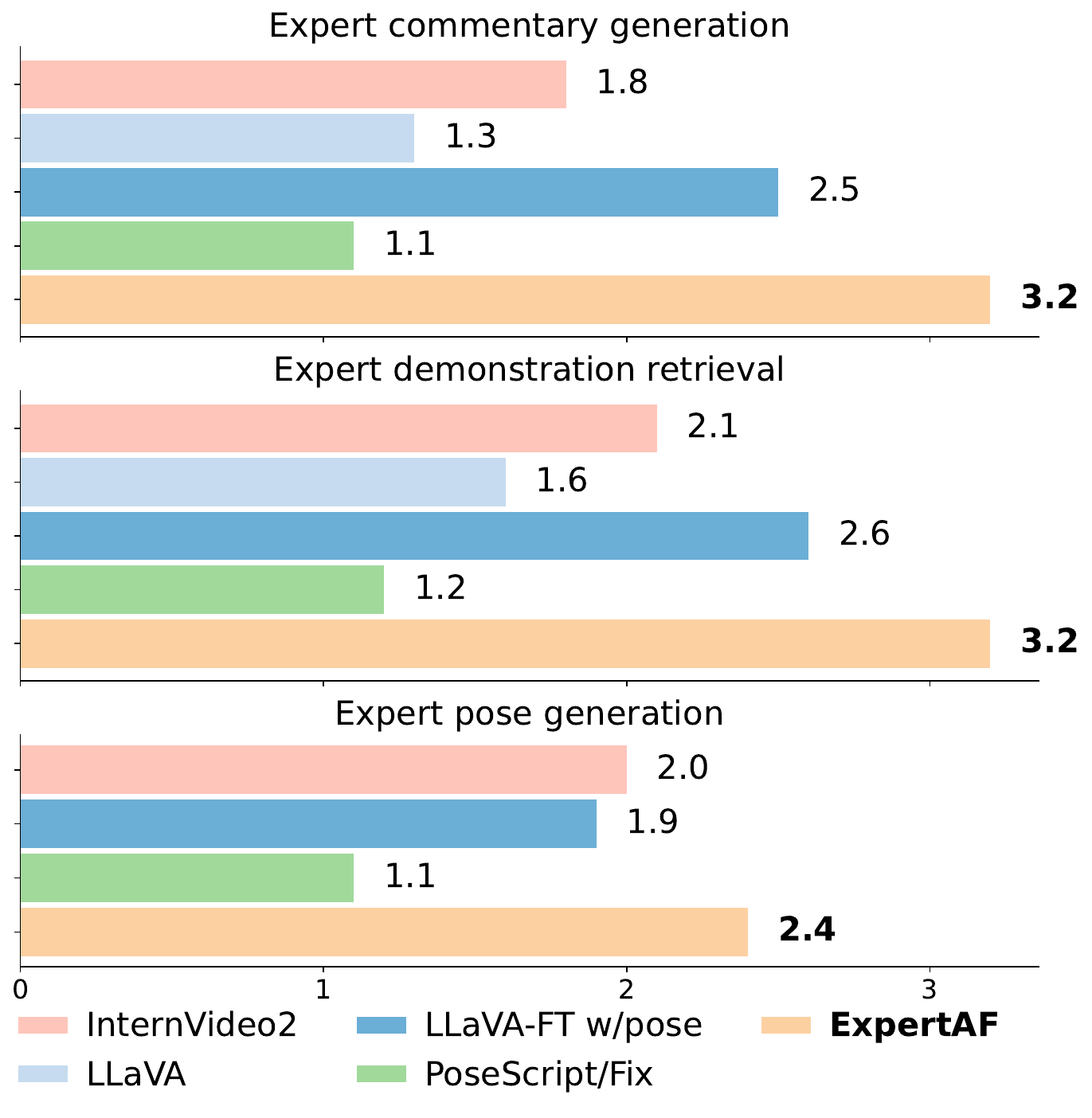}

    \end{minipage}
\vspace{-0.3cm}
    \captionof{table}{\textbf{Results on automatic metrics (left) and human evaluation (right).} We break down results for the three outputs---expert commentary generation, expert demonstration retrieval, and expert pose generation. Our method outperforms all baselines and prior work on all tasks. The last row ``w/ full-sup'' uses privileged input (the demo video $\bar{\mathcal{V}}$) at inference. (B@4: BLEU-4, M: Meteor, R-L: ROUGE-L F1, R: recall@50, medR: median rank, P: PA-MPJPE). For all metrics higher is better, except medR and PA-MPJPE ($\downarrow$). Our method is also rated higher by human raters on a Likert scale (min:1, max:4), compared to all the other methods %
    (right). See text for details. 
}
\label{tab:comprehensive_performance}
\vspace{-0.5cm}
\end{figure*}

\subsection{Implementation details}
\label{sec:implementation-details}

\textbf{Dataset and statistics.} Ego-Exo4D \cite{egoexo4d} contains 5,035 videos %
of participants doing activities across eight scenarios. %
We focus on three \emph{physical} scenarios---basketball, soccer, and rock climbing---though our model is in principle generalizable to other physical skills, without any task-specific design, unlike~\cite{workout-form-assessment,skating-eval}. 
 We use physical scenarios since the coaching feedback is groundable in the body regions, as opposed to procedural tasks like cooking, where the suggestions can be alternate ingredients or steps that are not visually present.
Details and statistics are in Supp.
Following the dataset creation process outlined in Sec.~\ref{sec:dataset-creation}, we obtain a dataset of 25,505 training and 1,272 testing tuples of $(\mathcal{V}, T, \bar{\mathcal{V}})$. We choose $k=5$ (train) and $k=1$ (test) for min-k choice of correct demonstrations per incorrect execution. To reiterate, we manually examine 
the test set for correct commentary summary and body region labeling.

\textbf{Network architecture.} 
The video model $f_V$ is an InternVideo2 \cite{internvideo2} encoder that provides strong visual representations. The pose encoder $f_P$ and decoder $f_P^{-1}$ are adapted from PCT \cite{pct}, which learns compositional tokens from human poses. Since the original training in \cite{pct} uses 2D poses, we adapt it to use 3-dimensions and retrain on the human poses in \cite{egoexo4d}.
Both the visual and pose mappers are low-parameter MLP layers \cite{improvedllava}. Finally, we use Llama 3-8B \cite{llama3modelcard} as the LLM multimodal encoder for sequence prediction. We finetune $\mathcal{F}_t$ and $\mathcal{F}_r$ for $5$ epochs with a learning rate of $5 \times 10^{-5}$.
Next, we modify the token dimension in $\mathcal{F}_g$ to accommodate pose tokens $\mathbf{p}'$, and hence, we fine-tune $\mathcal{L}_s$ when learning $\mathcal{F}_g$ with a learning rate of $5\times 10^{-6}$ for 5 epochs. Video ($f_V$) and pose models ($f_P$, $f_P^{-1}$) are kept frozen.
All the models are trained on 8 V100 32GB GPUs.

\section{Experiments and results}
\label{sec:results}

We first discuss the baselines and ablations, followed by the evaluation setup and results for the three outputs---expert commentary generation ($\mathcal{F}_t$), expert demonstration retrieval ($\mathcal{F}_r$), and pose demonstration generation ($\mathcal{F}_g$). We also show qualitative examples and discuss the limitations.

\noindent \textbf{Baselines.} We compare with the following baselines:
\begin{itemize}[leftmargin=0.2cm,itemsep=-0.0cm,parsep=0.1cm,topsep=-0.1cm]
    \item \textbf{InternVideo2-NN, InternVideo2-FT \cite{internvideo2}}: Given a query video $\mathcal{V}$, the nearest neighbor baseline (NN) finds the most similar video by InternVideo2 \cite{internvideo2} feature similarity in the training data and returns the corresponding commentary or demonstration, %
    for all tasks. The FT version finetunes the model to contrastively match the learner demonstration with the expert commentary/demonstration. %
    \item \textbf{VideoChat2 \cite{videochat,videochat2}, LLaVA \cite{llava}}: In these methods, we prompt SOTA video and image captioning models to generate commentary for an input demonstration. We use the log-likelihood loss to find the retrieved expert demonstration. These baselines evaluate if existing video captioning methods can provide expert feedback or retrieve expert demonstrations. We use the ego and ``best exo" frames, same as for our method. Note that %
    neither of these baselines are applicable for pose generation $\mathcal{F}_g$.
    \item \textbf{LLaVA-FT \cite{llava}, LLaVA-FT w/ pose \cite{llava}:} These baselines are based on the SOTA visual-language method LLaVA \cite{llava} but trained on our dataset with the same text model, i.e., Llama 3-8B \cite{llama3modelcard}, for an apples-to-apples comparison. The ``w/ pose'' variant also takes the 3D pose coordinates in text form as input.
    \item \textbf{PoseScript \cite{posescript}, PoseFix \cite{posefix}}: These two works enable pose-to-text and text-to-pose reasoning. The text generated or used for pose generation contains detail about the location of body parts, e.g., \emph{the hands are raised}, as opposed to expert commentary. Hence, they let us evaluate if pose description %
    is adequate for expert feedback. For pose generation, we evaluate if providing an expert commentary helps generate the desired expert demonstration. Both methods use SMPL \cite{smpl} pose and hence we convert 3D pose to SMPL and vice versa \cite{posetosmpl}, as needed.
\end{itemize}

\textbf{Ablations.} In addition to the strong baselines, we also compare the performance against ablations and variants of our design choices. %
\textbf{\modelname~ w/o pose} and \textbf{\modelname~ w/o video} evaluate the performance when only one modality is used. 
\textbf{\modelname~w/o alignment} quantifies the need for temporal alignment (Sec. \ref{sec:dataset-creation}), while \textbf{\modelname~ w/ global pose} evaluates if just providing the model with one correct pose per activity chosen based on expert commentary is enough (vs.~finding the closest correct pose sequence).  \textbf{\modelname~w/ full-sup} is a stronger variant with privileged input for inference---$\mathcal{F}_t$ uses $\mathcal{V}, \bar{\mathcal{V}}$ to predict $T$ and $\mathcal{F}_r$ and $\mathcal{F}_p$ uses both $\mathcal{V}$ and $T$. See Supp. for additional ablations of the choice of LLM $\mathcal{L}_s$, contribution of ego and exo videos, and joint training.

\textbf{Expert commentary generation.}
To evaluate the text commentary, we use standard metrics BLEU-4 \cite{bleu}, METEOR \cite{meteor} and ROUGE-L F1 \cite{rouge}, following prior work in evaluating text generation \cite{vast,captioning-metric} (higher is better).

\begin{figure*}
    \centering
    \includegraphics[width=\textwidth]{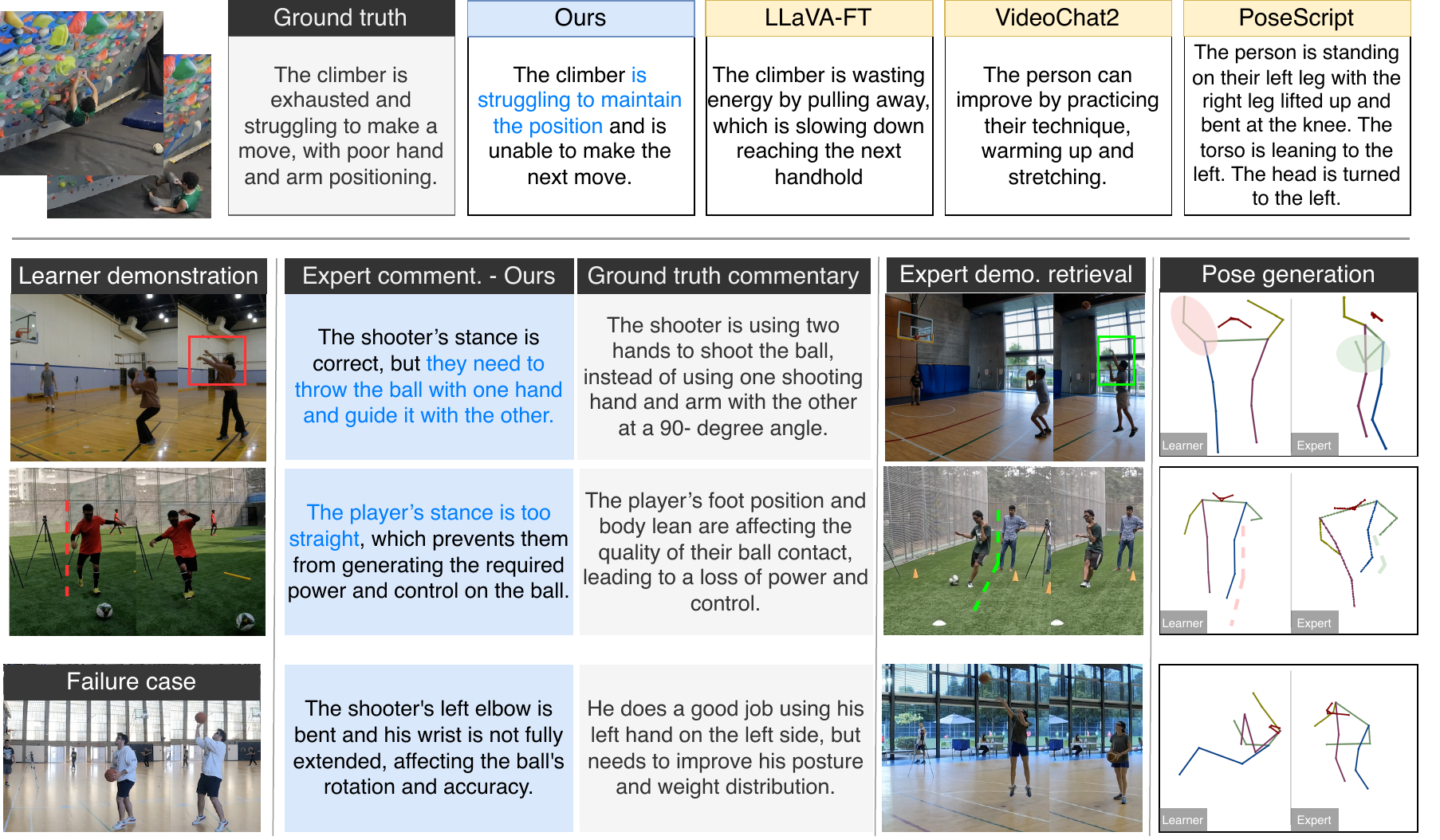}
    \caption{\textbf{Qualitative results.} (Top) Comparison of expert commentary generated by various baselines. (Second and third row) Examples of expert commentary generation, demonstration retrieval, and pose generation by our method. Notice the expert demonstration and pose generation corrects the mistake pointed out in the commentary, i.e., one hand is used to throw and the other to guide, and body position is improved to control better control and power. Colored text and marks (red for mistake, green for correction) are shown only for visualization. (Bottom) Failure cases.
    \textbf{See Supp. for video results.}
    } 
    \label{fig:qualitative}
    \vspace{-0.5cm}
\end{figure*}

Tab.~\ref{tab:comprehensive_performance} (left) shows the results. Our method outperforms all methods %
on all captioning metrics, with gains up to 3\% over the best baseline trained on the same dataset. All the gains are statistically significant using paired t-test with $p < 0.05$.
Fig. \ref{fig:qualitative} (top) shows commentary generation by various methods. Firstly, SOTA methods LLaVA \cite{llava} and VideoChat2 \cite{videochat2}, being captioning methods, %
yield a verbose description of the activity, rather than a critique. Fine-tuning LLaVA with our method improves the performance, but still it misses our use of expert demonstrations and pose sequences. Even using pose information as text only marginally improves ``LLaVA-FT w/ pose'', showing the effectiveness of our explicit pose encoding. Moreover, PoseScript \cite{posescript} generates pose descriptions like \emph{``head is turned to the left''} which is inadequate for feedback. Our results over the baselines shows the advantage of temporal alignment and a diverse expert demonstration set.

Our stronger variant ``w/ full-sup'' achieves an even better result by taking $\mathcal{V}$ and $\bar{\mathcal{V}}$ as inputs during inference. It is useful in cases where a learner has access to the expert demonstration but wants actionable feedback in language-form. Finally, Fig.~\ref{fig:qualitative} (bottom row) shows a failure case of commentary generation, showcasing the difficulty in pinpointing the exact mistake.

\textbf{Expert demonstration retrieval.}
For retrieval, we use standard retrieval metrics: recall@k and median rank. We set k=50 to account for the fact that multiple expert demonstrations could suitably correct the given learner demonstration. The retrieval set contains $1,272$ samples, the same as the test set size.  Higher recall and lower ranks are better.  %

Tab.~\ref{tab:comprehensive_performance} (middle) shows the results. Our method improves the median rank by 14 positions compared to the best baseline and a significantly better recall@50.
Our method and the baselines show a similar trend as with the expert commentary generation above, the stronger baseline w/ ``full-sup'' ($>$4\% better recall and 26 rank improvement) being crucial in cases where a learner has the commentary but wants a video exemplar to learn from.

\textbf{Expert pose generation.}
Tab.~\ref{tab:comprehensive_performance} (right column) shows the expert pose generation results, with error measured by PA-MPJPE~\cite{4dhumans} (mm). Our novel pose generation method performs better than learning 3D position using text in ``LLaVA-FT w/ pose''. %
Furthermore, PoseFix \cite{posefix} is designed to generate SMPL parameters based on a coarse modification text description. Our use of generation to output poses (vs.~retrieval) %
is advantageous when the candidate retrieval set does not contain the correct demonstration; see Supp. See Fig.~\ref{fig:qualitative} and Supp.~for qualitative examples including failure cases. The 3D pose sequences in the dataset are often auto-generated and have some noisy samples, discussed in Supp. Across the three tasks, our method is superior to its ablations.

\textbf{Human evaluation.}
For all the tasks, we also solicit %
human evaluation  (Tab.~\ref{tab:comprehensive_performance} right) to rate the quality of the generated text descriptions.  Five raters (per scenario) uninvolved with this project rate the quality of each generated output by scoring on a Likert scale from $1$ to $4$ (higher is better);details in Supp. We ensure that the raters have a basic knowledge of the scenario they are rating (basketball, soccer, rock climbing).

Across all three tasks, human raters score our method the highest. A significant gap of up to $0.7$ over the possible range $(1-4)$, as well as gains up to $3\times$, showcase \modelname's expert feedback quality.  That said, there is naturally room for improvement on this new task.  We find that while humans say our model excels on cases where the feedback is visually groundable, e.g. \emph{incorrect hand angle}, it tends to fall short when the feedback is not directly visible, e.g. \emph{the climber looks fatigued}.
The human evaluation complements the automatic metrics above, overcoming the limitation that a ``ground truth" commentary or a visual demonstration may capture only one of multiple possible errors in the learner's demonstration~\cite{gem,eval-issues}
(e.g., for a basketball shot, both the hand placement and the jump could be incorrect, but only one may be mentioned in the ground truth).
\section{Conclusion}
\label{sec:conclusion}

We proposed a novel task and method to generate expert commentary and demonstrations from a learner's video.  We develop %
a weakly-supervised training approach and benchmark for this problem. 
Our novel method fusing multi-modal inputs from learner and expert demonstrations together with expert commentary results in state-of-the-art performance in actionable feedback, and lays the groundwork for accessible, affordable, and actionable AI coaching applications in the future. 

\vspace{0.2cm}

\textbf{Acknowledgement.} UT Austin is supported in part by the IFML NSF AI Institute. Thanks to Fu-Jen Chu, Jing Huang and Xitong Yang for help with the Exo-Ego4D \cite{egoexo4d} pose extraction pipeline.

{
    \small
    \bibliographystyle{ieeenat_fullname}
    \bibliography{main}
}

\clearpage
\setcounter{page}{1}
\maketitlesupplementary

\appendix

\section{Supplementary video}
We provide a supplementary video containing an overview of the paper. The video contains details of the data collection approach, the method idea and finally, qualitative and quantitative results, \textbf{with video examples}.

\section{Expert feedback dataset}
\label{sec:supp-feedback-dataset}

In this work, we propose a novel expert feedback dataset that contains the tuple $(\mathcal{V}, \bar{\mathcal{V}}, T)$. The following subsections contain additional details about the specific prompt used, some limitations of the dataset and finally, some visualizations. Please also refer to the supplementary video for video demo of the dataset samples.

\subsection{Prompt for expert commentary classification and body region tagging}

In Sec. \ref{sec:dataset-creation}, we outline the process of obtaining a weakly-supervised training data from commentaries in Ego-Exo4D~\cite{egoexo4d}. The first step involves summarizing the expert commentary into a single actionable feedback, followed by tagging the body region to whether the execution is correct, incorrect or not mentioned at all. We use the following prompt to ask Llama 3-70B \cite{llama3modelcard}:

\begin{tcolorbox}[breakable, boxrule=0.2mm]
\textbf{System:} Llama3 default system prompt.

 \textbf{User:} The following is an expert commentary about a person playing Basketball. Give a one sentence summary of the expert feedback and then mention which body parts out of Head, Shoulder, Hands, Arms, Legs, Jump needs improvement and which ones are good execution (you can choose multiple body parts). Here is the expert's commentary: 
 
 He came down on one foot that time. You want to make sure you come down on two feet. Right now he's putting a little bit too much pressure and stress. Then on top of that, his left knee is locked, which could easily cause some hypertension. So be very aware and careful of your landing.

 \textbf{Assistant:} One sentence summary: He came down on one foot and his left knee is locked, which could cause some hypertension. \textbackslash n Needs improvement parts: Legs, Jump. \textbackslash n Good execution parts: None.

 \textbf{User:} Let's take a look at the placement of the shooter's left guy hand. You can see that it's in a really good position on the left side of the ball. If you take a look at the shooting hand, his right hand is underneath the ball with his right index finger in the middle section of the ball. His right index finger is in the middle section of the ball. This is good positioning of both his right shooting hand and his left guy hand on the ball.
 
 \textbf{Assistant:} One sentence summary: Shooter's hand is in a really good position on the left side of the ball. \textbackslash n Needs improvement parts: None. \textbackslash n Good execution parts: Hands.

 \textbf{User:} $<$new\_question$>$

 \textbf{Assistant:}     
\end{tcolorbox}

This example prompt is the template for basketball, and we follow a similar process for other scenarios. To ease the generation process, we provide one incorrect and one correct execution example, shown above. Subsequently, we parse the outputs and discard the commentaries that cannot be parsed. Most of the discarded commentary contains no information to reliably classify it, as done above. As included in the prompt above, we use the following body regions for all the scenarios for a coarse classification:

\begin{tcolorbox}[breakable, boxrule=0.2mm]
Head, Shoulder, Hands, Arms, Legs, Jump
\end{tcolorbox}

\subsection{Expert feedback classification examples}

\begin{figure*}
    \centering
    \includegraphics[width=\textwidth]{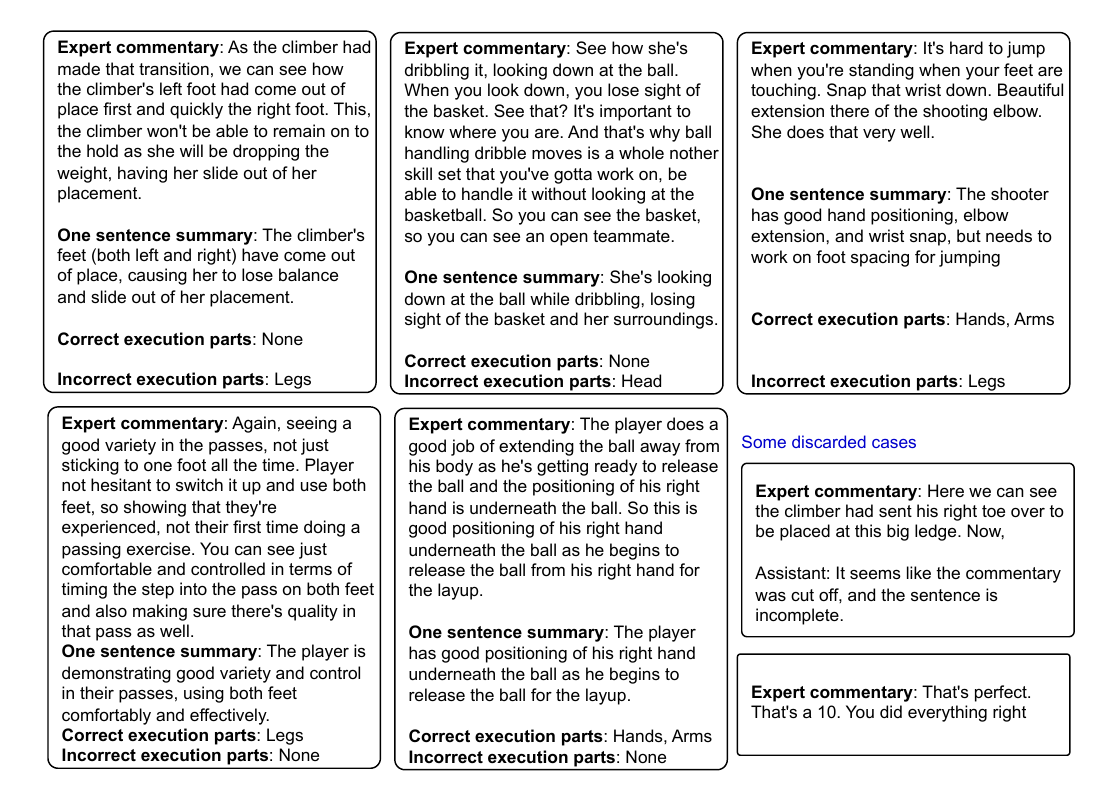}
    \caption{\textbf{Expert commentary classification examples.} Llama3 correctly generates a one-sentence summary of the expert commentary, along with a tagging of the body region with incorrect and correct execution. The last example shows two discarded examples.}
    \label{fig:expert-comm-examples}
\end{figure*}

Fig. \ref{fig:expert-comm-examples} shows some examples of expert commentary preprocessing. The above described prompt results in these samples. We obtain a summarized expert commentary, along with label of correct and incorrect execution for every body region. Recall that the expert commentaries are obtained by converting expert speech to text using ASR. Consequently, some samples contain incomplete information and, hence, discarded (see bottom right in Fig. \ref{fig:expert-comm-examples}).

\subsection{Visualization of the dataset}

\begin{figure*}
    \centering
    \includegraphics[width=\textwidth]{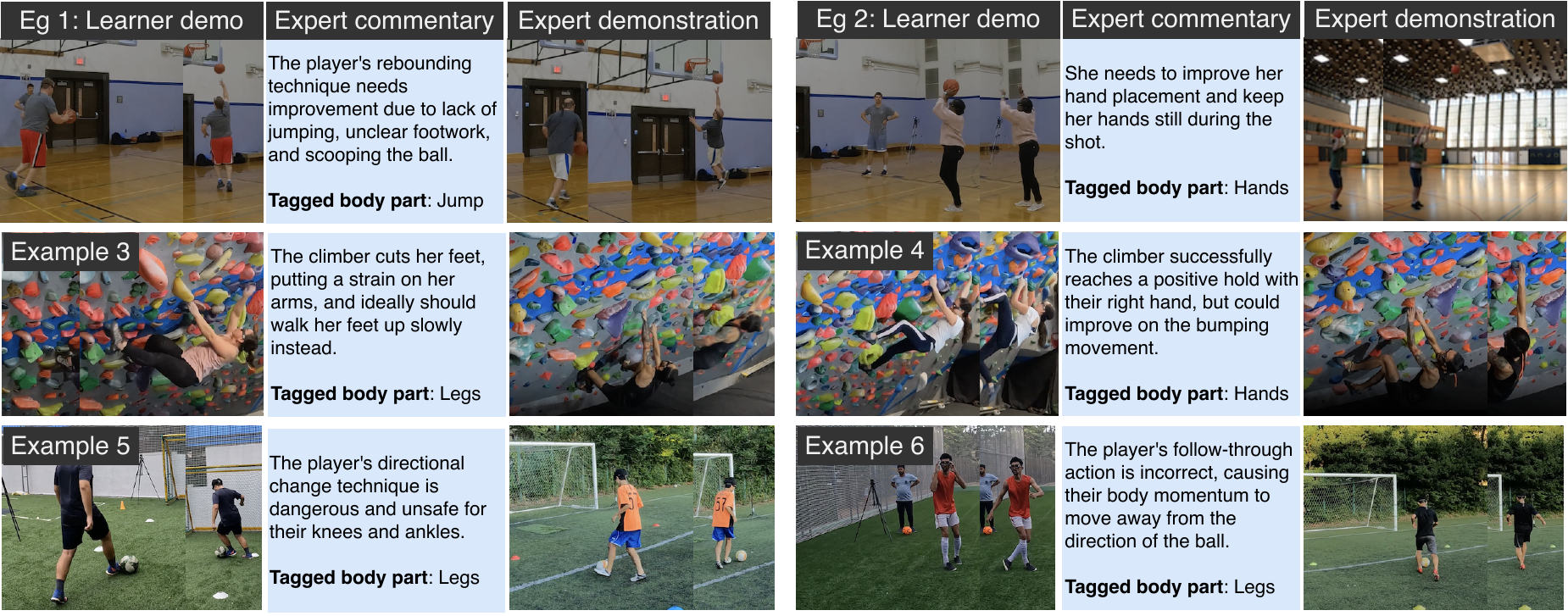}
    \caption{\textbf{Examples from the expert feedback dataset.} Some examples of the expert feedback tuples $(\mathcal{V}, T, \bar{\mathcal{V}})$ generated by our approach. Notice how the expert demonstration corrects the errors in the learner demonstration. For example, the top left video shows the person attempting a shot without jumping, as noted in the expert commentary. This error is corrected in the expert demonstration, where the person jumps correctly when shooting.}
    \label{fig:supp-data-viz}
\end{figure*}

\begin{figure*}[t]
    \centering
    \includegraphics[width=\textwidth]{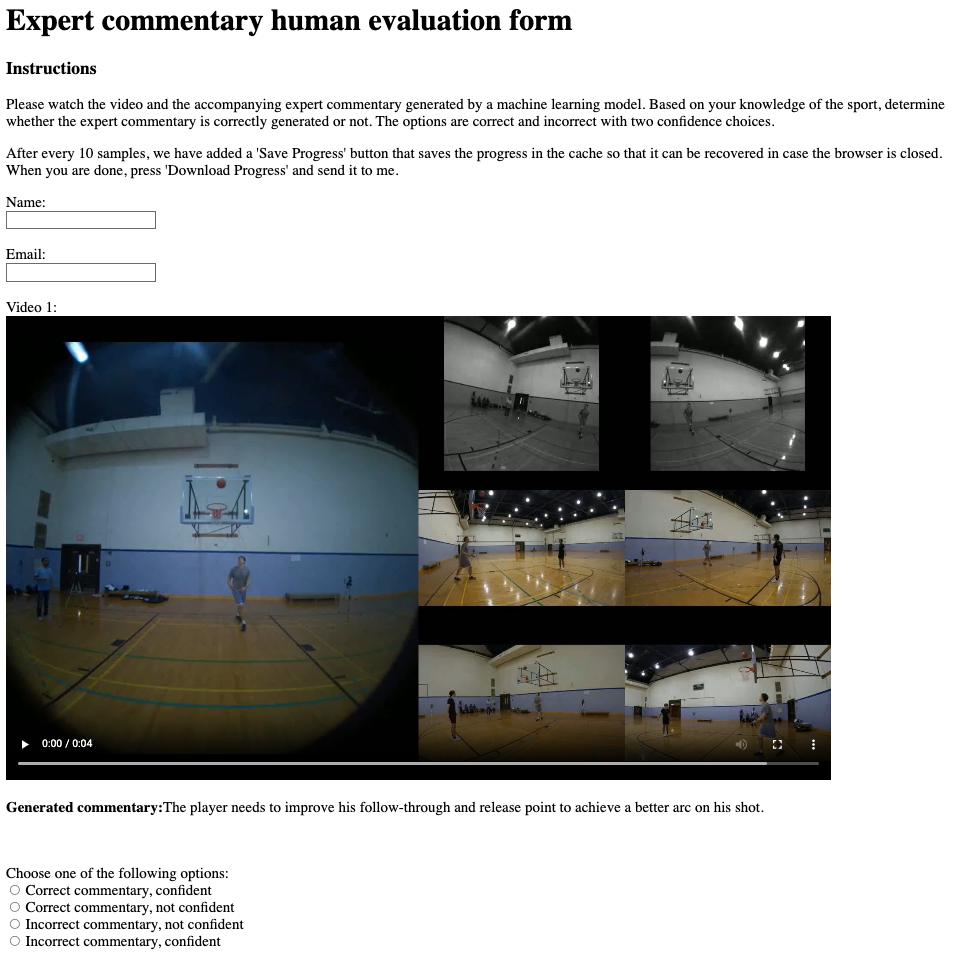}
    \caption{\textbf{Human evaluation interface.} A screenshot of the human evaluation interface.}
    \label{fig:human-eval-interface}
\end{figure*}

Fig. \ref{fig:supp-data-viz} shows six samples from the weakly-supervised expert feedback dataset. See the video examples in the attached supplementary video. Each example contains the tuple $(\mathcal{V}, T, \bar{\mathcal{V}})$. All the learner demonstrations have errors that are pointed out in the expert commentary (center). Finally, the error is corrected in the expert demonstration. For example, the jump is incorrect (Example 1) or the follow-through after shooting is incorrect (Example 6). The same is corrected in the expert demonstration. Note that all the videos in the dataset are accompanied by 3D poses.

\begin{figure*}
    \centering
    \includegraphics[width=\textwidth]{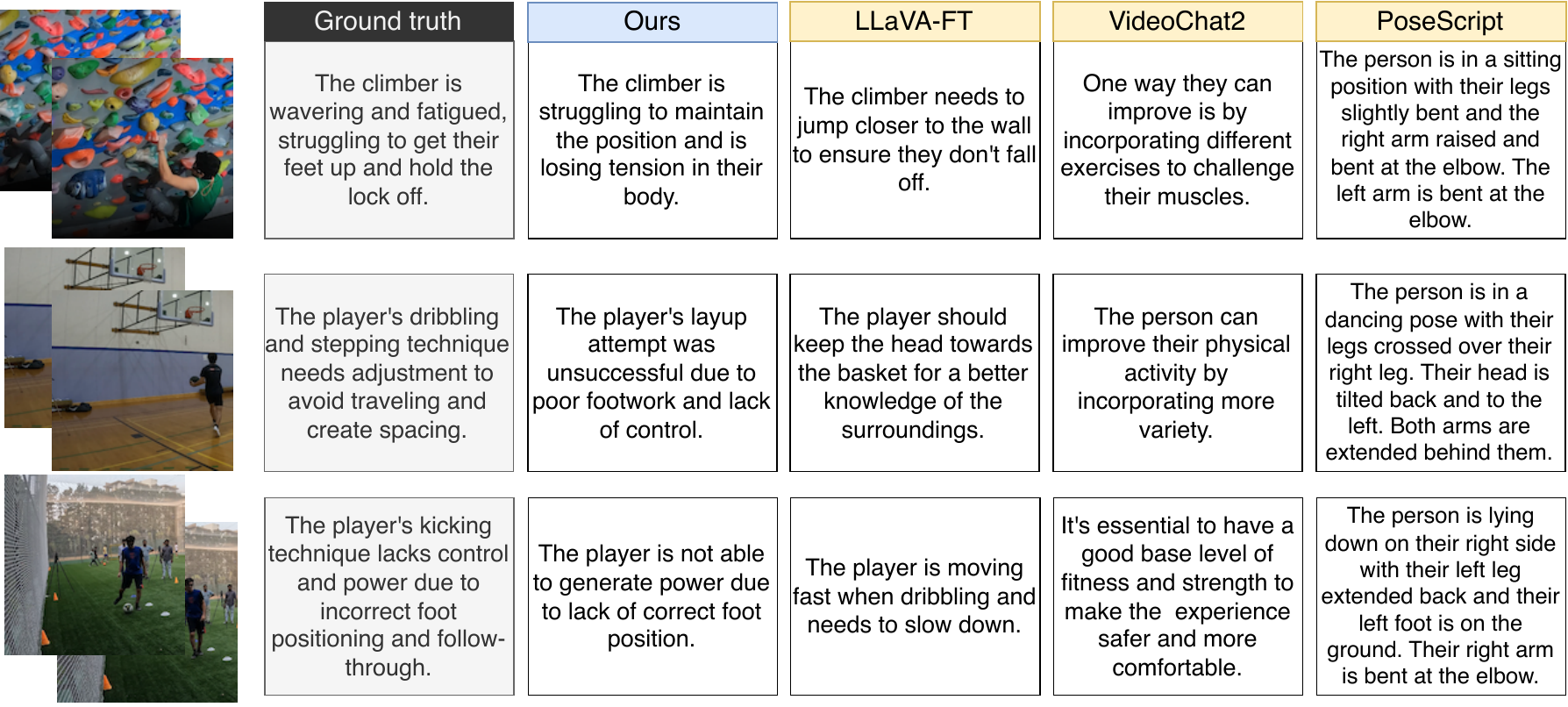}
    \includegraphics[width=\textwidth]{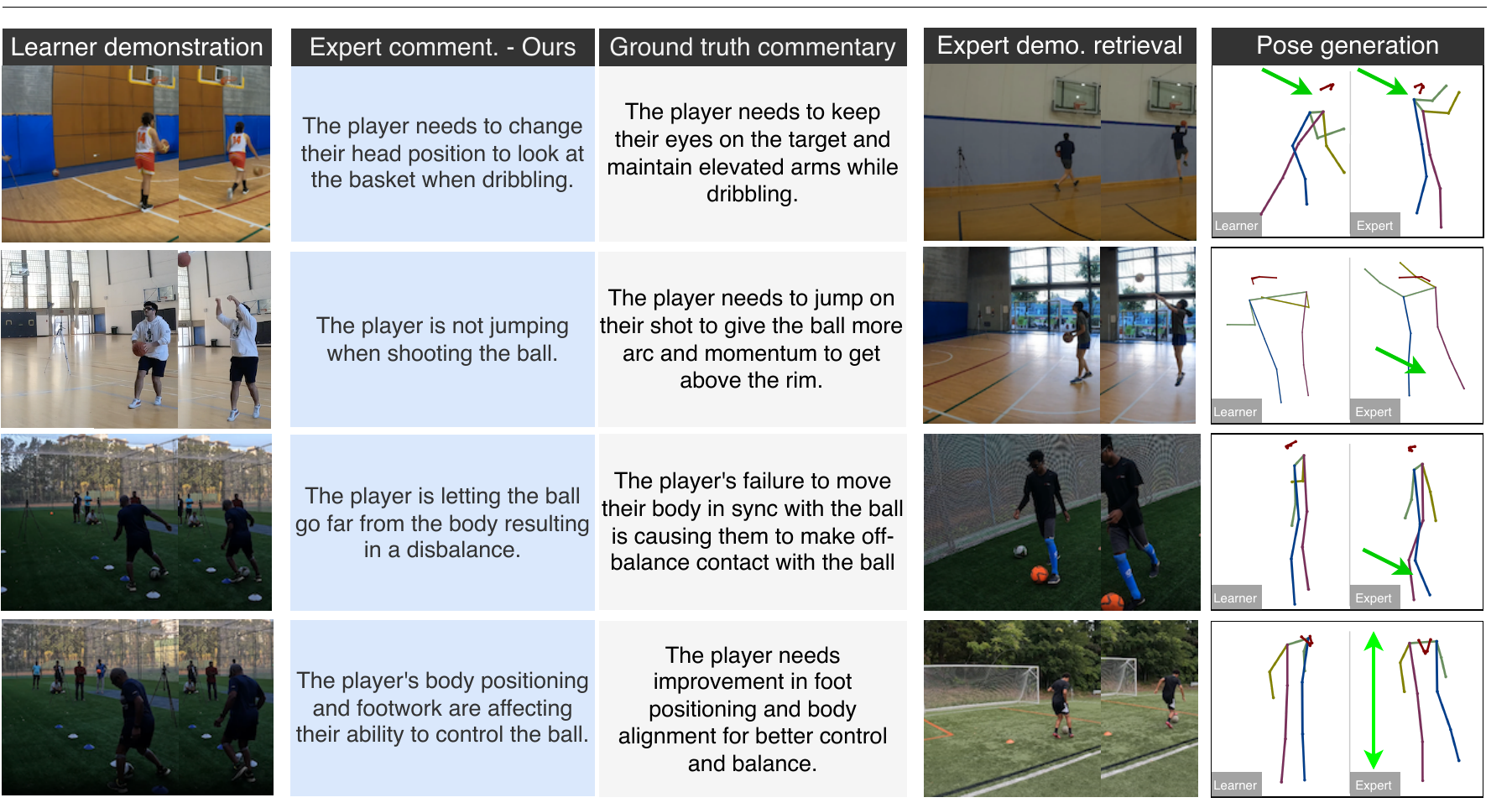}
    \caption{\textbf{Additional results visualization.} (Top) Comparison of expert commentary generated by various baselines. (Bottom) Examples of expert commentary generation, demonstration retrieval, and pose generation by our method.}
    \label{fig:supp-additional-viz}
    \vspace{-0.5cm}
\end{figure*}

\subsection{Dataset statistics}
As noted in Sec. \ref{sec:implementation-details}, we use Ego-Exo4D \cite{egoexo4d} for our experiments. We choose the physical scenarios in the dataset---basketball, rock climbing and soccer. These physical scenarios contain 34092 expert commentaries. The expert commentary classification and body region localization results in labeling $16791$ commentaries as having incorrect executions of at least one body region and $20946$ commentaries with at least one good execution. Some commentaries do not contain enough information for classification and are discarded. There are $2073$ distinct participants in the scenarios of interest and those are classified into four categories---novice ($340$), early expert ($420$), intermediate expert ($642$) and late expert ($671$). The created expert feedback dataset contains $25505$ training and $1272$ testing samples. Each training and testing clip is $4$ seconds long, which is the typical duration of an execution in the selected physical scenarios.

\subsection{Manual verification of the test set}

We verify the automatically curated test set to obtain a clean subset of $1,272$ samples. The verification process firsts check if the expert commentary is correctly summarized into a single sentence, along with a correct assignment of the body region. Next, we examine the tuple $(\mathcal{V}, T, \bar{\mathcal{V}})$ for correctness. We check if the feedback in $T$ is incorporated in $\bar{\mathcal{V}}$. We only keep the segment if we are certain that the expert demonstration corrects the mistake in $T$. Overall, only $5.5\%$ of the samples were discarded---showcasing the quality of the automated pipeline.

\section{Additional implementation details}

The demonstration clips are 4 seconds long and we sample frames at 32 fps.
The InternVideo2 \cite{internvideo2} video encoder $f_V$ takes in 8 frames and thus, generates 4 features per second (totalling 16 features). We concatenate the ego and exo features to create 32 input tokens for each demonstration. In the expert demonstration retrieval training, we use the log likelihood loss as the relevance score, lower is better.

\section{Additional ablations}

In addition to the ablations discussed in Sec. \ref{sec:results}, we also evaluate the choice of the LLM $\mathcal{L}_s$, the contribution of ego and exo videos, and the joint training with all scenarios. Table~\ref{tab:ablation} summarizes the performance for all these ablations. We discuss each of them below:

\textbf{Effect of the choice of the LLM.} We observe that the performance increases with the relative strength of the LLM. This experiment suggests that the expert actionable feedback will futher improve with advancements in these strong language models. Notably, our method will still be useful to learn the fine-grained differences between learner and expert demonstrations, and to provide actionable expert feedback.

\textbf{Effect of separate training.} We jointly train all the three scenarios---\emph{basketball}, \emph{soccer} and \emph{rock climbing}. We observe that training all scenarios separately does not improve the performance. The performance with separate training remains lower due to no cross-scenario learning and lower generalizability.

\textbf{Contribution of ego and ego views.} We observe a better performance for exocentric-only ablation. This result is expected since exo view captures the body pose more accurately. However, if we use pose + egocentric views (`w/ ego $+$ pose'), the performance increases since the body pose compensates for the missing exocentric video.

\textbf{Generation vs retrieval when the expert demonstration is missing in the candidate set.} To evaluate the quality of the generation in the absence of correct demonstration in the retrieval candidate set, we remove the 50 closest samples with the ground truth from the retrieval set, to simulate missing correct demonstrations. Our retrieval method finds the best remaining demonstration, and extracts pose. See Tab.~\ref{tab:ablation} ``ret. w/o GT''. The PA-MPJPE error is 160 mm which is worse than the error in the third task.

\begin{table}[t]\footnotesize
        \centering
\begin{tabular}{L{2.0cm}C{0.50cm}C{0.50cm}C{0.50cm}C{0.50cm}C{0.50cm}C{0.50cm}}
\toprule
&  \multicolumn{3}{c}{\textbf{Comm. Gen.}} & \multicolumn{2}{c}{\textbf{D.R.}}  & \multicolumn{1}{c}{\textbf{P.}} \\
\cmidrule(lr){2-4} \cmidrule(lr){5-6} \cmidrule(lr){7-7}
\textbf{ExpertAF} &  B@4 & M & RL  & R & mR & P~$\downarrow$  \\
\midrule
w/ Llama3.2 1B & 42.8  & 48.5   & 53.0  & 17.4  & 177 &  153  \\
w/ Llama3.2 3B & 43.6  & 49.1 & 53.8  & 18.3  & 168  & 145   \\
\midrule
w/ sep. train & 43.2  & 48.0  & 52.6  & 15.6  & 176  & 155  \\
\midrule
ret. w/o GT & ---  & ---  & ---  & ---  & ---  & 160  \\
\midrule
w/ ego only & 40.0  & 45.7  & 50.1  & 15.8  & 194  & ---  \\
w/ exo only & 44.2  & 49.0  & 54.1  & 18.3  & 169  & ---  \\
w/ ego $+$ pose & 44.4  & 49.2  & 54.0  & 18.3  & 164 & 144  \\
\midrule
Ours & \textbf{44.9} & \textbf{49.6} & \textbf{54.6} & \textbf{19.1} & \textbf{158} & \textbf{135}  \\
\bottomrule
\end{tabular}
    \caption{\textbf{Results of additional ablations.} Our method outperforms all ablations. See the text for discussion. (Comm. Gen.: Expert commentary generation, D.R.: Expert demonstration retrieval, P.: Expert pose generation).}
    \label{tab:ablation}
\end{table}

\section{Additional result visualization}

Fig.~\ref{fig:supp-additional-viz} contains additional result visualization. We see that our method generates better expert commentary compared to all baselines. Similarly, our expert demonstration retrieval and pose generation correct the mistakes in the input demonstration. \textbf{See the video visualization in the attached supplementary video.}

\section{Limitations}

We observe the following limitations:

\emph{Incorrect 3D pose for some samples.} Recall that the 3D pose sequence in \cite{egoexo4d} is calculated by triangulating the position from various exo cameras placed around the subject. Thus, the 3D pose sequence is an \emph{auto-ground truth}. Therefore, some samples have reconstruction error. However, this issue is beyond the scope of our work.

\emph{Variable expert commentary granularity.} There are many experts annotating the dataset. Therefore, some experts give fine-grained feedback about legs, hands etc. whereas some experts provide coarse instructions like incorrect posture. A consistent granularity is desirable for a better training, but we do not ignore any sample for a more diverse expert commentary generation.

\section{Human evaluation details}
\label{sec:supp-human-eval}

\subsection{Evaluation setup}

We obtain a random subset of 250 samples (out of 1272) for human evaluation. For each scenario, we ask 5 raters to evaluate all the methods. As mentioned in Sec. \ref{sec:results}, we group the methods for efficient rating, scoring 5 groups of baselines for each of the three scenarios. The raters are appropriately compensated. To reiterate, we ensure all the raters for each scenario have $2+$ years of experience of doing that physical scenario.

\subsection{Evaluation interface and instructions}

Fig. \ref{fig:human-eval-interface} shows the human evaluation interface, along with the instructions. We have similar interface for expert commentary retrieval, and pose generation. The sampled are randomized for each method to avoid bias while annotating.

\clearpage

\end{document}